\documentclass[lettersize,journal]{IEEEtran}
\usepackage{amsmath,amsfonts}
\usepackage{algorithm}
\usepackage{array}
\usepackage[caption=false,font=normalsize,labelfont=sf,textfont=sf]{subfig}
\usepackage{textcomp}
\usepackage{stfloats}
\usepackage{url}
\usepackage{verbatim}
\usepackage{graphicx}
\usepackage{cite}
\usepackage{multirow}
\usepackage{booktabs}
\usepackage{algpseudocode}
\usepackage{array}
\usepackage{xcolor}

\begin{document}

\title{UniqueSplat: View-conditioned 3D Gaussian Splatting for Generalizable 3D Reconstruction}

\author{Haixu Song, Xiaoke Yang, Shengjun Zhang, Jiwen Lu,~\IEEEmembership{Fellow,~IEEE}, Yueqi Duan,~\IEEEmembership{Member,~IEEE}
\thanks{Haixu Song, Xiaoke Yang, Shengjun Zhang, Jiwen Lu and Yueqi Duan are with Tsinghua University, China (e-mail: shx22@mails.tsinghua.edu.cn; yangxk22@mails.tsinghua.edu.cn; zhangsj23@mails.tsinghua.edu.cn; lujiwen@tsinghua.edu.cn; duanyueqi@tsinghua.edu.cn).

Corresponding author: Yueqi Duan}
}

\markboth{Journal of \LaTeX\ Class Files,~Vol.~14, No.~8, August~2021}%
{Shell \MakeLowercase{\textit{et al.}}: A Sample Article Using IEEEtran.cls for IEEE Journals}

\IEEEpubid{0000--0000/00\$00.00~\copyright~2021 IEEE}

\maketitle

\begin{abstract}
In this paper, we propose UniqueSplat, a view-conditioned feed-forward 3D Gaussian Splatting model to reconstruct customized 3D radiance fields for each view query. 
Existing feed-forward methods such as pixelSplat and MVSplat aim to generate fixed Gaussians across all views of each scene by minimizing the error between rendered views and ground-truth images.
However, such fixed Gaussians generally render images from all views and lack the ability to adapt to specific viewpoints, as they do not incorporate target view information when predicting Gaussians.
To address this, our UniqueSplat learns the view-conditioned information as a prior and incorporates this knowledge into network parameters, so that Gaussians are dynamically adjusted in accordance with different views.
Specifically, we propose a two-branch view-conditioned hyperNetwork to simultaneously learn view-agnostic embeddings and view-specific knowledge, which not only explores the shareable knowledge from various views, but also adapts the model to specific views at test time. 
Extensive experiments on widely-used datasets including RealEstate10K, ACID and DTU demonstrate the superiority of UniqueSplat over the state-of-the-art methods.
Moreover, UniqueSplat encouragingly outperforms existing methods in cross-dataset evaluation, showing its notable generalization ability.
\end{abstract}

\section{Introduction}
 \begin{figure*}
    \centering
    \includegraphics[width=\linewidth]{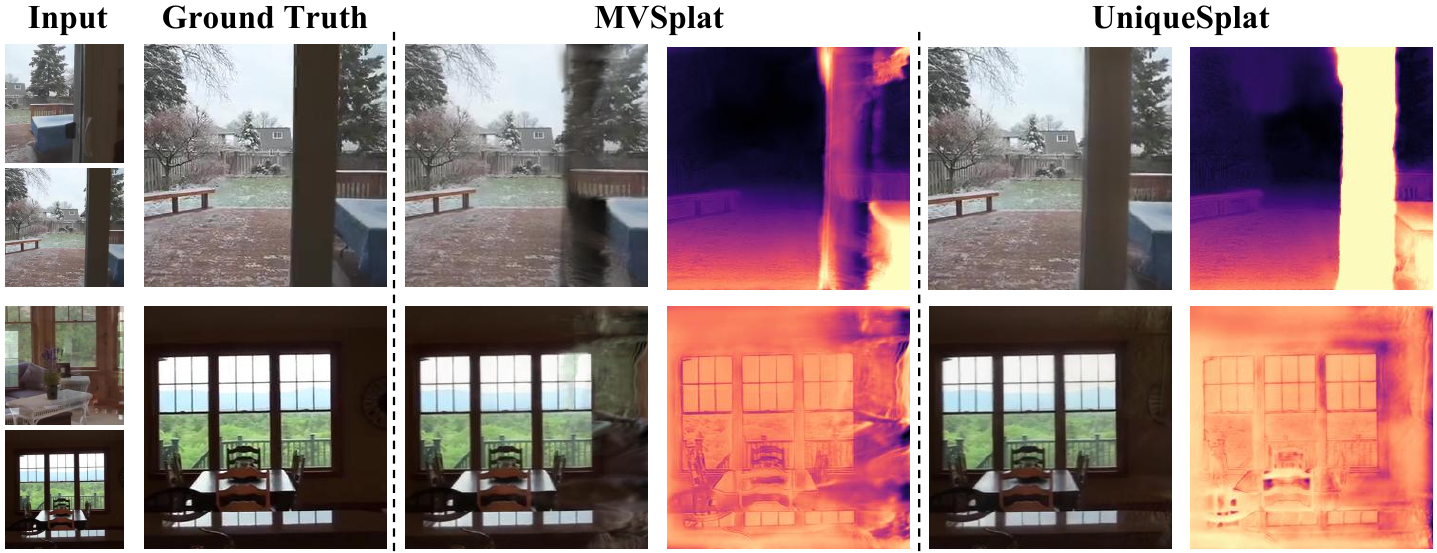}
    \caption{Qualitative comparison between MVSplat~\cite{chen2024mvsplat} and UniqueSplat. UniqueSplat incorporates view-conditioned information as a prior to achieve superior quality in novel view synthesis and produce high-quality depth maps, while MVSplat presents issues such as cracks, blurs, and artifacts.}
    \label{fig:teaser}
\end{figure*}

\IEEEPARstart{N}{ovel} view synthesis is a crucial problem in computer vision due to its broad applications, such as virtual reality~\cite{zheng2024gps}, robotics~\cite{jiang2023h} and autonomous driving~\cite{tonderski2024neurad}. Traditional approaches represent scenes using point clouds~\cite{wang2023joint}, meshes~\cite{cui2022vidsfm,li2023nr}, or a 4D light field function~\cite{jin2022occlusion}.
Neural implicit representations have achieved great success in recent years, while such methods typically require substantial computational resources for both training and rendering, which limits their efficiency and scalability~\cite{liu2020neural,barron2021mip,fridovich2022plenoxels,muller2022instant,chen2024learning}.
Recently, 3D Gaussian Splatting (3DGS)~\cite{3DGS2023ToG} has drawn increasing attention which represents the scene with a set of 3D Gaussian primitives to enable real-time rendering. 
This approach leverages rasterization-based rendering to avoid dense point querying, thereby maintaining both high efficiency and quality.

Nevertheless, 3DGS requires substantial images to optimize each specific scene, which presents challenges for generalizable 3D reconstruction across diverse scenes~\cite{yan2024multi,yu2024mip}. 
To address this, several generalizable methods have been proposed to improve generalizability and transferability from sparse view images~\cite{szymanowicz2024splatter,chen2024mvsplat,zhang2025transplat,zhang2024GGN}.
These methods utilize feed-forward networks to predict the parameters of pixel-aligned Gaussians in single forward pass, enhancing both efficiency and applicability in reconstructing unseen scenes. For instance, pixelSplat~\cite{charatan2024pixelsplat} employs epipolar transformers to unproject depth maps from corresponding views to predict the parameters of 3D Gaussians, while MVSplat~\cite{chen2024mvsplat} leverages cost volume representation to avoid the noisy artifacts.

While these methods have achieved encouraging performance, they focus on learning a feed-forward model by minimizing the loss between rendered views and ground-truth images, aiming to predict fixed Gaussians with the minimal average error across all views for each scene.
However, while such fixed Gaussians generally render images from mulitple views, they cannot specialize to a particular viewpoint, since the target view information is not incorporated into the Gaussian prediction process.
As illustrated in Figure~\ref{fig:teaser}, MVSplat exhibits issues such as cracks, blurs and artifacts when synthesizing novel views.
\IEEEpubidadjcol
In this paper, we propose UniqueSplat, a view-conditioned feed-forward 3D Gaussian Splatting model that reconstructs customized 3D radiance fields for each view query. 
Unlike existing methods that predict fixed Gaussians for rendering all views of a scene, our UniqueSplat learns view-conditioned information as a prior and integrates the knowledge into the network parameters to customize 3D radiance fields, enabling the Gaussian fields to dynamically adjust for each specific view. 
Specifically, instead of directly relying on the input views, we decompose the view-conditioned knowledge into two components for each view query: view-agnostic information, which remains consistent across views, and view-specific information, which varies for each view query in the scene.
Therefore, we design a view-conditioned hypernetwork consisting of two branches: a view-agnostic branch and a view-specific branch, in contrast to prior hypernetwork-based NVS methods that adopt only a single branch.
The view-agnostic branch maintains learnable embeddings which are continuously updated during training to abstract the characteristics across various views of the training scenes. 
Since the view-agnostic branch generates a fixed set of parameters to predict Gaussians across multiple views during testing, we introduce the view-specific branch as a complementary to dynamically adjust the predictor for each given view.
To achieve this, we construct the feature representation via combining the features of input sparse-view images for each view query, and then project it into an embedding to incorporate view-specific knowledge. Finally, we fuse the weights generated by both branches to customize Gaussians for each individual view.

We summarize our key contributions as follows:
\begin{itemize}
    \item We propose UniqueSplat, a method for novel view synthesis from sparse view images. While existing methods leverage fixed Gaussians to render all views, we are the first to customize Gaussians which dynamically adjust according to the view query.
    \item We design a dual-branch view-conditioned hypernetwork consisting of a view-agnostic branch and a view-specific branch to capture shareable knowledge and view-specific information, unlike prior NVS methods that employ hypernetworks to directly generate weights solely from input signals. 
    \item We conduct extensive experiments on widely-used benchmark datasets, including RealEstate10K, ACID and DTU. Our method achieves state-of-the-art results in both intra-test and cross-test settings. 
\end{itemize}

\section{Related Works}
\label{sec:related_works}

\subsection{Single-scene Novel View Synthesis}
Neural rendering~\cite{mildenhall2020nerf,barron2021mip} has attracted significant interest in recent years due to its impressive capability to synthesize high-quality novel views from 3D scenes. 
However, despite their remarkable visual fidelity, these methods often suffer from high computational costs. This issue arises because rendering a single image typically requires querying a neural field multiple times per ray, resulting in slow performance, particularly for real-time applications.
To mitigate these costs, subsequent studies have introduced discrete data structures to accelerate the rendering process~\cite{chen2022tensorf,fridovich2022plenoxels,liu2020neural,muller2022instant,zhang2023fast}. While these methods successfully accelerate the rendering process to some extent, they still face limitations in achieving real-time performance, particularly when rendering at high resolutions.
More recently, 3D Gaussian Splatting (3DGS)~\cite{3DGS2023ToG} has emerged as a promising technique in novel view synthesis. 
By employing an efficient rasterization-based splatting approach, 3DGS leverages a set of 3D Gaussian primitives to represent a scene. This novel technique offers considerable advancements in both rendering speed and quality, making it a compelling alternative to earlier methods that rely on dense neural representations. 
Despite these significant advantages, 3DGS can still produce artifacts during the splatting process, motivating ongoing research efforts to improve the quality and realism of rendered views~\cite{yan2024multi, jiang2024gaussianshader, liang2024gs, yu2024mip}.
Another key challenge associated with 3DGS is its high storage requirement. To represent a single scene accurately, 3DGS often necessitates millions of parameters, leading to substantial memory and storage demands.
To tackle this issue, several research efforts~\cite{lu2024scaffold, navaneet2023compact3d, girish2023eagles} focus on reducing memory usage to enable real-time rendering without compromising on quality of the rendered scenes. 
Additionally, some works~\cite{zhu2024fsgs, xiong2023sparsegs} aim to minimize the number of input images required for scene reconstruction, further enhancing the efficiency of 3DGS. 
Moreover, several studies~\cite{wu20244d, yang2023real, xie2024physgaussian}  have explored extending the application of 3D Gaussians into the realm of 4D scenarios by incorporating dynamic elements that align with real-world physical behavior.
Besides, recent works have also incorporated physical priors or novel neural representations to further improve rendering quality.
3DGS-Enhancer~\cite{liu20243dgs} leverages 2D video diffusion priors to address the challenging 3D view-consistency problem by reformulating it as achieving temporal consistency in video generation. Mvpgs~\cite{xu2024mvpgs} exploits multi-view priors to enhance the quality of geometric initialization for 3DGS and further introduces multiple constraints to improve rendering quality. LITA-GS~\cite{zhou2025lita} incorporates an illumination-invariant physical prior and develops a lighting-agnostic rendering strategy to achieve high-quality, normally exposed representations under adverse illumination conditions. In addition, Chung et al.~\cite{chung2024depth} propose a depth-guided Gaussian Splatting optimization strategy that mitigates overfitting by refining scene optimization with depth priors.
However, these methods primarily focus on optimizing a specific scene. When applied to a new scene, they require additional time for optimization and lack generalization ability.

\subsection{Generalizable Novel View Synthesis}
Unlike optimization-based methods, which typically require dozens of minutes to train a separate model for each scene, feed-forward networks directly synthesize novel views in a single inference, requiring substantially less time compared with optimization-based methods.
This results in superior cross-dataset generalizability and faster rendering speeds. In recent years, implicit representations have parameterized fully connected neural networks to represent complex 3D scenes with global and local codes~\cite{yu2021pixelnerf,jang2021codenerf,rematas2021sharf}.
Some approaches enhance the quality of neural radiance field reconstruction via constructing cost volumes at pre-defined reference viewpoints~\cite{chen2021mvsnerf,johari2022geonerf,liu2022neural}, while others perform independent ray-based rendering~\cite{chibane2021stereo,suhail2022generalizable,varma2022attention,wang2021ibrnet}.
Additionally, there are methods specifically designed for large baseline settings~\cite{du2023learning,sajjadi2022scene}, which focus on reconstructing 3D scenes from a wider range of camera positions.
Despite significant advancements in MLP-based representations, these methods are often slow to render, limiting their potential for real-time applications. To address this, a variety of faster representations have been developed, including voxel grid representations~\cite{sun2022direct,szymanowicz2023viewset} and triplane representations~\cite{anciukevivcius2023renderdiffusion,gu2023nerfdiff}.
Voxel grids significantly accelerate rendering by directly encoding opacities and colors, although they scale poorly with resolution.
The triplane representation, on the other hand, enables view-space reconstruction and provides performance suitable for single-view reconstruction.
In recent years, there has been a surge of interest in generalizable 3D Gaussian Splatting (3DGS) methods for reconstructing 3D scenes from sparse views. 
For instance, Splatter Image~\cite{szymanowicz2024splatter} and GPS-Gaussian~\cite{zheng2024gps} focus on regressing pixel-aligned Gaussian parameters to reconstruct specific objects or humans rather than complex scenes. 
In contrast, pixelSplat~\cite{charatan2024pixelsplat} leverages sparse view inputs to predict Gaussian parameters by utilizing epipolar geometry and depth estimation. 
MVSplat~\cite{chen2024mvsplat} introduces a cost volume representation to directly predict depth from cross-view features, significantly improving geometric quality. 
LatentSplat~\cite{wewer2024latentsplat} addresses reconstruction challenges in object-centric scenes with 360° views by incorporating variational Gaussians, improving both accuracy and completeness.
TranSplat~\cite{zhang2025transplat} applies depth-aware deformable matching and monocular depth priors to enhance reconstruction quality.
However, these feed-forward methods generate fixed Gaussians across all views of a scene, failing to adapt to each individual view. To address this limitation, we propose UniqueSplat, which incorporates view-conditioned information to customize Gaussians for each view query, resulting in more realistic and high-quality images.

\subsection{HypernetWorks}

Hypernetworks~\cite{ha2016hypernetworks} are a class of neural networks designed to generate weights for another neural network, known as the primary network, based on a specific input embedding. 
This input embedding can be derived from various types of data, allowing the hypernetwork to conditionally adjust the weights of the primary network depending on the context. 
Hypernetworks have emerged as a powerful deep learning technique, enabling greater adaptability, dynamism and information sharing~\cite{chauhan2024brief}.
They have been integrated into models for tasks such as image recognition, semantic segmentation, and neural architecture search, where they replace convolutional or linear layers~\cite{nirkin2021hyperseg, brock2018smash}.
In the context of implicit neural representations, hypernetworks have proven particularly useful for generating task-conditioned networks that can adapt to new data instances, making implicit representations more flexible for tasks where the scene or data is subject to change. Recently, hypernetworks have been successfully applied to sound~\cite{szatkowski2022hypersound}, representation compression~\cite{gordon2024d} and novel view synthesis~\cite{chen2022transformers,wu2023hyperinr,sen2023hyp}.
However, such hypernetwork-based NVS methods primarily focus on generating implicit neural representation directly from input signals. In contrast, our proposed UniqueSplat employs a view-conditioned hypernetwork to explicitly model Gaussian representation, predicting dynamic Gaussians conditioned on the query view. By adopting a dual-branch design comprised of a view-specific branch and a view-agnostic branch, our model effectively disentangles scene knowledge, thereby achieving more consistent and adaptable novel view synthesis.
\section{Method}
\label{sec:method}

In this section, we first provide an overview of the proposed UniqueSplat in Section~\ref{sec:overview}. We then describe the view-conditioned hypernetwork in Section~\ref{sec:hypernetwork} and introduce the primary network in Section~\ref{sec:primary}.

\subsection{Overview} 
\label{sec:overview} 
\begin{figure}
    \centering
    \includegraphics[width=1\linewidth]{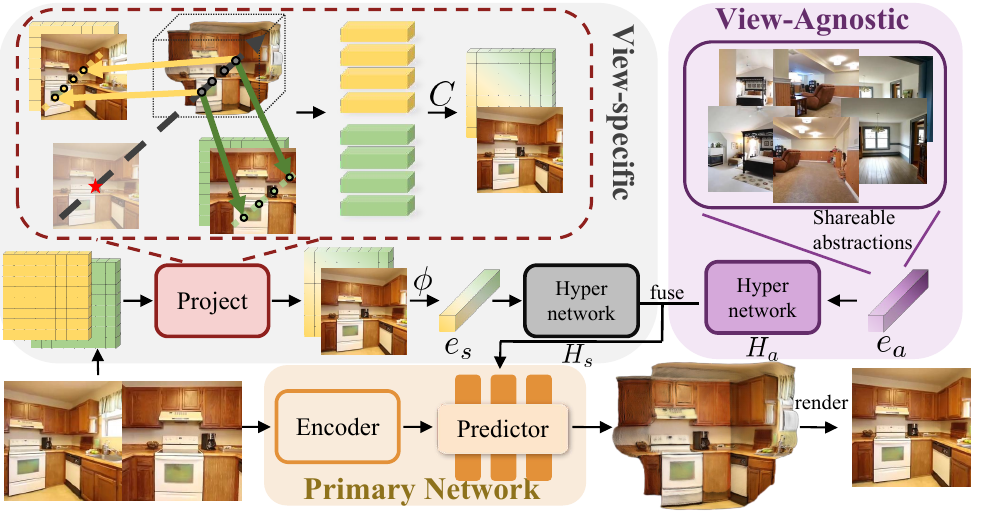}
    \caption{
    Overview of our UniqueSplat framework, consisting of two key components: the primary network and the view-conditioned hypernetwork, which includes a view-specific branch and a view-agnostic branch. 
    The view-specific branch leverages the projection matrix to project features from input views (\textcolor{yellow}{yellow} and \textcolor{green}{green}) to the target viewpoint, incorporating the knowledge of specific view. 
    It then maps the target view feature into an embedding and calculates a corresponding weight.
    Meanwhile, the view-agnostic branch extracts shareable knowledge from multiple views of the training scenes and generates another weight via a hypernetwork. 
    Finally, the weights are fused and serve as the weights of the target layers, ensuring the integration of both view-specific and view-agnostic information.}
    \label{fig:pipeline}
\end{figure}

Figure~\ref{fig:pipeline} provides an overview of our proposed method, which consists of a primary network and a view-conditioned hypernetwork.The primary network incorporates an encoder and a predictor, while the hypernetwork is further divided into two complementary branches: view-agnostic and view-specific. This dual structure allows the model to combine both general and distinctive information for multiple views.

Given $K$ sparse-view images $I_c = \{{I}_{c}^{i}\}_{i=1}^{K}$, their corresponding camera projection matrices $P_c = \{{P}_{c}^{i}\}_{i=1}^{K}$ and $N$ corresponding camera projection matrices of target views $P_t=\{{P}_{t}^{i}\}_{i=1}^{N}$ , our objective is to render novel views utilizing a feed-forward neural network architecture. The encoder $E_p$ in primary network processes each image to produce a feature volume represented as $F_p = \{{F}_{p}^{i}\}_{i=1}^{K}$. This encoder captures the intrinsic information within individual views and facilitates the exchange of features across different views. By integrating features from multiple viewpoints, the encoder allows the model to better understand the underlying structure and context of the scene, enhancing the robustness of the extracted representations.

To effectively incorporate view information into our model, we employ a view-conditioned hypernetwork to obtain parameters for the predictor dynamically. Unlike conventional methods that maintain fixed layers during inference, our hypernetwork obtains parameters according to the input images. Specifically, the view-specific branch extracts specific view information by mapping the feature projected from feature of input views to generate an embedding $e_{s}$, which is subsequently processed by the hypernetwork $H_{s}$ to produce the corresponding weights $W_{s}$. Concurrently, learnable view-agnostic embeddings $e_{a}$ are processed by the view-agnostic hypernetwork $H_{a}$ to yield weights $W_{a}$.
We then fuse these weights into weight $W_{f}$ and apply it to the target layers of the predictor. By incorporating these weights, the predictor in primary network effectively integrates the view prior, allowing it to predict Gaussian parameters including the center $\mu$, opacity $\sum$, covariance $\sigma$, and color $c$. Finally, novel views are rendered from the predicted 3D Gaussians through a splatting operation, enabling the generation of high-fidelity images from various viewpoints.  

\subsection{View-conditioned Hypernetwork}
\label{sec:hypernetwork}
In our approach, the view-conditioned hypernetwork injects view priors into the target layers for predicting Gaussian parameters. It consists of two branches: a view-agnostic branch and a view-specific branch.
\subsubsection{View-agnostic Branch}
The view-agnostic branch of our model is specifically designed to extract shareable information from diverse input views. This enables the model to maintain a comprehensive understanding of the scene that is not tied to any specific viewpoint.

To achieve this, we introduce learnable embeddings, denoted as $e_{a}$, which capture generalizable features which are applicable across multiple viewpoints. We then employ a hypernetwork $H_{a}$ to project them into a weight matrix $W_{a}$ as follows:

\begin{equation} 
W_a = H_{a}(e_{a}), \quad {W_a} \in \mathbb{R}^{C_{out} \times C_{in}},
\end{equation}
where $C_{in}$ and $C_{out}$ denote the input and output dimensions of the target convolution layers respectively.
The weight matrix \( W_{a} \) encodes a set of embeddings derived from the diverse input views of training scenes, which captures information in a manner which is not dependent on a particular viewpoint.

Once the view-agnostic branch generates the weight matrix $W_{a}$, we integrate this matrix into the subsequent layers of our primary network. This incorporation allows us to inject the extracted view-agnostic knowledge into the model, allowing the model to leverage this shared information to improve the synthesis quality across different viewpoints.

\subsubsection{View-specific Branch}

The view-specific branch is a key component of our model, designed to learn a view-specific embedding vector \(e_{s}\) for each view, which matches the dimensionality of the view-agnostic embedding \(e_{a}\). 
Unlike \(e_{a}\), which remains fixed during testing, the view-specific embedding \(e_{s}\) varies across different views, enabling the network to incorporate the information of specific view for accurate rendering.

To accomplish this, we first employ a backbone encoder \(E_H\) to extract per-view features $F_c=\{{F}_{c}^{i}\}_{i=1}^{K}$ from a series of images $I_c = \{{I}_{c}^{i}\}_{i=1}^{K}$ where each image \(I_{c}^{i}\) represents a distinct viewpoint of the same scene. 
The feature extraction process can be represented as follows:
\begin{equation}
    {F}_{c}^{i} = E_H(I_c^i), {I}_c^i \in \mathbb{R}^{H \times W \times 3}.
\end{equation}

Following feature extraction, we compute the corresponding projection matrices \(P_c=\{P_{c}^{i}\}_{i=1}^{K}\) for each input image, as well as the projection matrices \(P_t=\{P_{t}^{i}\}_{i=1}^{N}\) for the target views. Each projection matrix is formulated as:

\begin{equation}
    {P}^{i} = {K}^{i}[{R}^{i}|{t}^{i}],
\end{equation}
where \({K}^{i}\) represents the intrinsic parameters of the camera, while \({R}^{i}\) and \({t}^{i}\) denote the rotation and translation matrices, respectively. 

To effectively incorporate view prior information, we utilize the target view projection matrix to construct a feature volume by projecting the source features to align with the target view as follows:
\begin{equation}
    F_{t}^i=\textrm{Proj}(F_{c}, P_{c}, P_{t}^i),
    \label{equ:proj}
\end{equation}
where $F_c$ and $P_c $ denote the feature maps and projection matrices of the input views, and ${P}_{t}^i$ represents the projection matrix of \(i\)-th target view.

For clarity, we derive it in detail, focusing on a single target view as an example.
Given the projection matrix \(P_t = K_t[R_t | t_t]\) for the target view, we generate a ray for each pixel \((u_t, v_t)\) in the target view.
First, we convert the pixel coordinates \((u_t, v_t)\) to normalized camera coordinates by inverting the intrinsic matrix \(K_t\):
\begin{equation}
    \mathbf{p}_t = K_t^{-1} \begin{bmatrix} u_t \\ v_t \\ 1 \end{bmatrix},
\end{equation}
where \(\mathbf{p}_t = [x_t, y_t, 1]^\top\) represents the point in normalized camera coordinates.

Using the extrinsic parameters \([R_t | t_t]\) of the target view, we map \(\mathbf{p}_t\) from the camera coordinate system to the world coordinate system:
\begin{equation}
  \mathbf{r} = R_t^\top \mathbf{p}_t, \quad \mathbf{o} = t_t,
\end{equation}
where \(\mathbf{r}\) is the ray direction, and \(\mathbf{o}\) is the ray's origin (i.e., the camera's position in world coordinates).

Thus, the ray in world coordinates is represented as:
\begin{equation}
\mathbf{x}(\lambda) = \mathbf{o} + \lambda \mathbf{r}, \quad \lambda \in \mathbb{R}.
\end{equation}

Next, given the projection matrix \(P_c = K_c[R_c | t_c]\) for the source views, we project the ray onto the source views to obtain the corresponding epipolar lines:
\begin{equation}
\mathbf{p}_c(\lambda) = P_c \mathbf{x}(\lambda) = P_c (\mathbf{o} + \lambda \mathbf{r}).
\end{equation}

This projection results in epipolar lines in the source views, which can be expressed as a linear function of the pixel coordinates \(\mathbf{p}_t\):
\begin{equation}
\mathbf{L} = \mathbf{A} \mathbf{p}_t + \mathbf{b},
\end{equation}
where \(\mathbf{A} = K_c R_c R_t^\top\) represents the relative rotation between the target and source views, and \(\mathbf{b} = K_c (R_c t_t + t_c)\) encodes the relative translation. The epipolar lines \(\mathbf{L}\) correspond to the projection of the ray onto the image planes of both input views.

Subsequently, we uniformly sample \(n\) points along \(F_c\) between the near and far planes. 
We then calculate a correlation matrix to evaluate the relationship between the projected feature from input views for each pixel. We define a function \(C\) to compute the correlation matrix of two matrices \((M_1, M_2)\) as follows:

\begin{equation}
\label{equ:correlation_func}
    \begin{aligned}
        C_{ij}(M_1, M_2) = \frac{(M_1^i, M_2^j)}{\| M_1^i \| \| M_2^j \|},
    \end{aligned}
\end{equation}
where \(M_i^j\) represents the \(j\)-th row of matrix \(M_i\), \((\cdot)\) indicates the inner product operation, and \(\|\cdot\|\) denotes the norm of a vector. This correlation matrix enables us to effectively fuse the projected features by aggregating them according to their correlation.

Following the fusion of the projected features, we map each target view feature into an embedding that incorporates the specific knowledge of the current view:

\begin{equation}
    e_{s} = \phi (F_{t}).
\end{equation}

We then leverage this embedding to obtain the weight matrix \(W_{s}\) through a hypernetwork \(H_{s}\):

\begin{equation}
\label{equ:fuse_weight}
    \Delta W = H_{s}(e_{s}), \quad \Delta{W} \in \mathbb{R}^{C_{out} \times C_{in}}.
\end{equation}

Since the dimensions of \(W_{s}\) and \(W_{a}\) are identical, we directly fuse these weights to obtain a weight \(W\) which combines both information from distinct branches.

\begin{equation}
    {W}_H = {W} + \Delta{W}, \quad {W_H} \in \mathbb{R}^{C_{out} \times C_{in}},
\end{equation}
where \({W} \) also retains the same dimensions as the target layer. This fusion process enables the model to integrate both general and specific view information, thereby enhancing the overall rendering performance and ensuring high fidelity in the rendered novel views.

\subsection{Primary Network}
\label{sec:primary}
Similar to MVSplat~\cite{chen2024mvsplat}, the primary network consists of an encoder \(E_p\) and a predictor. The encoder processes the input images to extract feature volumes \(F_p = \{F_p^i\}_{i=1}^K\), where each \(F_p^i\) represents the feature map extracted from the input images.
This transformation is mathematically described as:
\begin{equation}
    F_p = E_{\text{p}}(I_c, P_c),
\end{equation}
where \(I_c\) and \(P_c\) represent the input images and their corresponding camera parameters, respectively. 

Next, we employ a depth predictor to estimate the parameters of a set of 3D Gaussian primitives including the mean \(\mu\), covariance \(\Sigma\), opacity \(\alpha\), and color \(c\), which collectively define the properties of each primitive in 3D space. 
In order to ensure that the model effectively incorporates view-conditioned information, we strategically integrate hypernetworks into the convolutional layers of the predictor. This integration allows the model to dynamically adjust based on the specific view query of the scene, thereby improving the accuracy of the predictions.

After obtaining the fused weights via the view-conditioned hypernetwork as in (\ref{equ:fuse_weight}), we reshape these weights to match the dimensions required by the convolutional layers. This ensures that the weights are compatible with the architecture of the predictor.
These weights are then used by the predictor to generate the Gaussian parameters, which are essential for rendering realistic images from the target viewpoint. The prediction process is formalized as:
\begin{equation}
\mu, \Sigma, \alpha, c = D_{\text{injected}}(F_p),
\end{equation}
where \(\mu\), \(\Sigma\), \(\alpha\), and \(c\) represent the center, covariance, opacity, and color of the Gaussian representation, respectively. 

Once the predictor generates the Gaussian parameters, we leverage these predicted Gaussians to render images from the target view. This rendering process is defined as:
\begin{equation}
    I_t = R(G, P),
\end{equation}
where \(I_t\) represents the rendered images, \(G\) denotes the Gaussians generated by the predictor, and \(P\) is the projection matrices of rendered views.

\begin{algorithm}[t]
\caption{UniqueSplat}\label{alg:uniquesplat}
\renewcommand{\algorithmicrequire}{\textbf{Input:}}
\renewcommand{\algorithmicensure}{\textbf{Output:}}
\begin{algorithmic}
\Require Input images $I_{c}$, camera projection matrix of input views $P_{c}$, camera projection matrix of target views $P_{t}$
\Ensure Target views $I_{t}$
\State $W=H_{a}(e_{a})$
\For{$i\leftarrow 1$ to $K$}
    \State $F_{c}^{i}=E_{H}(I_{c}^{i})$
\EndFor
\For{$i\leftarrow 1$ to $N$}
    \State $F_{t}^{i}=\textrm{Proj}(F_{c}, P_{c}, P_{t}^{i})$
\EndFor
\State $e_{s}=\phi(F_{t})$
\State $\Delta W=H_{s}(e_{s})$
\State $W_H=W + \Delta W$
\State $F_{p}=E_{p}(I_{c}, P_{c})$
\State $\mu, \Sigma, \alpha, c = D_{\textrm{injected}}(F_{p})$
\State $I_{t}=R(G, P_t)$
\end{algorithmic}
\end{algorithm}

\subsection{Optimization}

To train our full model, we utilize ground truth target RGB images to provide supervision during the learning process. In order to ensure a balance between pixel-level accuracy and perceptual quality, we carefully design the training loss as a linear combination of two distinct loss functions: the Mean Squared Error (MSE) loss and the Learned Perceptual Image Patch Similarity (LPIPS) loss. This dual loss approach ensures that the model is capable of capturing both fine-grained pixel-wise details as well as high-level perceptual features. The overall loss is defined as:

\begin{equation}
    L = \lambda_1 L_{mse}(\Bar{I}_t, I_t) + \lambda_2 L_{lpips}(\Bar{I}_t, I_t),
\end{equation}
where \(\Bar{I}_t\) represents the ground truth target image, and \(I_t\) represents the novel view synthesized by the model.

By optimizing this loss function, our model learns to minimize pixel-wise discrepancies while enhancing perceptual quality, leading to high-fidelity image synthesis.

\subsection{Discussion}

\subsubsection{Comparison with Ray-Decoupled Image-Based Rendering Methods}
Several prior works in image-based rendering (IBR) adopt strategies that decouple rays, enabling independent ray-based rendering. 
Specifically, IBRNet~\cite{wang2021ibrnet} employs a ray transformer to aggregate information along each ray within the NeRF decoder, while SRF~\cite{chibane2021stereo} predicts color and density for each 3D point based on an encoding of its stereo correspondence, implicitly learned from an ensemble of pairwise similarities. Differently, GNT~\cite{varma2022attention} introduces a view transformer to aggregate multi-view image features and a ray transformer to predict target colors, directly regressing pixel colors for view synthesis without relying on NeRF’s volume rendering. GPNR~\cite{suhail2022generalizable} leverages 2D transformer blocks to implicitly capture correspondence by aggregating features along epipolar lines across multiple views, followed by pixel color prediction using the fused features. In contrast, MatchNeRF~\cite{chen2025explicit} explicitly computes feature correspondences via cosine similarity for the NeRF conditional inputs. Overall, these methods conduct independent ray-based rendering and do not explicitly model 3D context across rays. In contrast, our method leverages view-conditioned information as a prior, incorporating it into the network to predict explicit pixel-aligned Gaussians for the input views. By explicitly encoding scene geometry, this approach enables more accurate reconstruction of scene structures, as also reflected in the qualitative results.

\subsubsection{Comparison with Hypernetwork-Based Novel View Synthesis Methods}
Hypernetworks have recently been applied in novel view synthesis to predict implicit neural representation (INR) weights. For instance, Trans-INR~\cite{chen2022transformers} employs a Transformer hypernetwork to infer the entire set of INR weights, removing the single-vector bottleneck and avoiding grid-based representation or gradient computation. HyperINR~\cite{wu2023hyperinr} directly predicts compact INR weights via a hypernetwork, achieving state-of-the-art inference performance. HyP-NeRF~\cite{sen2023hyp} estimates both the weights and multi-resolution hash encodings, yielding quality improvements compared to estimating only the weights of a NeRF.
In summary, existing NVS methods primarily focus on directly generating the weights of implicit neural representations from the input signals. 
In contrast, our approach emphasizes explicit Gaussian representations while incorporating view priors rather than relying solely on the raw input. 
While prior hypernetwork-based methods typically generate shared weights across different views within the same scene, our method proposes a view-conditioned hypernetwork that predicts distinct weights conditioned on the query view.
To the best of our knowledge, UniqueSplat is the first feed-forward Gaussian Splatting model to predict dynamic Gaussians across views.
To enable this adaptability, we design a view-conditioned hypernetwork that leverages hypernetworks to adapt across different view queries. Unlike prior approaches that adopt a single-branch design, our hypernetwork consists of two complementary branches: (1) a view-agnostic branch, which captures information shared across diverse views, and (2) a view-specific branch, which captures information unique to each specific view by fusing input-view features with the target view matrix. This dual-branch design is motivated by the observation that scene knowledge can be naturally decomposed into view-specific and view-agnostic components.

\section{Experiment}
\label{sec:experiment}
\begin{table*}[t]
    \centering
    \caption{Quantitative comparison on RealEstate10K~\cite{zhou2018stereo} and ACID~\cite{liu2021infinite} benchmarks. We report the average results of PSNR, SSIM and LPISP on all test scenes. Our method outperforms baselines on all evaluation metrics.}
    \normalsize
    \setlength{\tabcolsep}{5mm}{\begin{tabular}{ccccccc}
        \toprule
        \multirow{2}{*}{Method} & \multicolumn{3}{c}{RealEstate10K~\cite{zhou2018stereo}} & \multicolumn{3}{c}{ACID~\cite{liu2021infinite}} \\ 
        \cmidrule(lr){2-4} \cmidrule(lr){5-7}
        & PSNR↑ & SSIM↑ & LPIPS↓ & PSNR↑ & SSIM↑ & LPIPS↓ \\
        \cmidrule(lr){1-7}
        pixelNeRF~\cite{yu2021pixelnerf} & 20.43 & 0.589 & 0.550 & 20.97 & 0.547 & 0.533 \\
        GPNR~\cite{suhail2022generalizable} & 24.11 & 0.793 & 0.255 & 25.28 & 0.764 & 0.332 \\
        AttnRend~\cite{du2023learning} & 24.78 & 0.820 & 0.213 & 26.88 & 0.799 & 0.218 \\
        MuRF~\cite{xu2024murf} & 26.10 & 0.858 & 0.143 & 28.09 & 0.841 & 0.155 \\
        pixelSplat~\cite{charatan2024pixelsplat}   & 25.89 & 0.858 & 0.142 & 28.14 & 0.839 & 0.150 \\
        MVSplat~\cite{chen2024mvsplat}  & 26.39 & 0.869 & 0.128 & 28.25 & 0.843 & 0.144 \\
        TranSplat~\cite{zhang2025transplat}   & 26.69 & 0.875 & 0.125 & 28.35 & 0.845 & 0.143 \\
        Ours  & \textbf{27.28} & \textbf{0.881} & \textbf{0.119} & \textbf{29.06} & \textbf{0.855} & \textbf{0.134} \\
        \bottomrule
    \end{tabular}}
    \label{tab:intratest}
\end{table*}

\begin{table*}[]
    \centering
    \caption{Qualitative comparison for cross-dataset generalization. All models are trained on RealEstate10K~\cite{zhou2018stereo} and tested on ACID~\cite{liu2021infinite} and DTU~\cite{jensen2014large} without any fine-tuning.}
    \normalsize
    \setlength{\tabcolsep}{5mm}{\begin{tabular}{ccccccc}
        \toprule
        \multirow{2}{*}{Method} & \multicolumn{3}{c}{ACID~\cite{liu2021infinite}} & \multicolumn{3}{c}{DTU~\cite{jensen2014large}} \\ 
        \cmidrule(lr){2-4} \cmidrule(lr){5-7}
        & PSNR↑ & SSIM↑ & LPIPS↓ & PSNR↑ & SSIM↑ & LPIPS↓ \\
        \cmidrule(lr){1-7}
        pixelSplat~\cite{charatan2024pixelsplat}  & 27.64 & 0.830 & 0.160 & 12.89 & 0.382 & 0.560 \\
        MVSplat~\cite{chen2024mvsplat}  & 28.15 & 0.841 & 0.147 & 13.94 & 0.473 & 0.385 \\
        TranSplat~\cite{zhang2025transplat}  & 28.17 & 0.842 & 0.146 & 14.93 & 0.531 & 0.326 \\
        Ours & \textbf{28.63} & \textbf{0.848} & \textbf{0.142} & \textbf{16.01} & \textbf{0.665} & \textbf{0.285} \\
        \bottomrule
    \end{tabular}}
    \label{tab:crosstest}
\end{table*}

\begin{figure*}
    \centering
    \includegraphics[width=\linewidth]{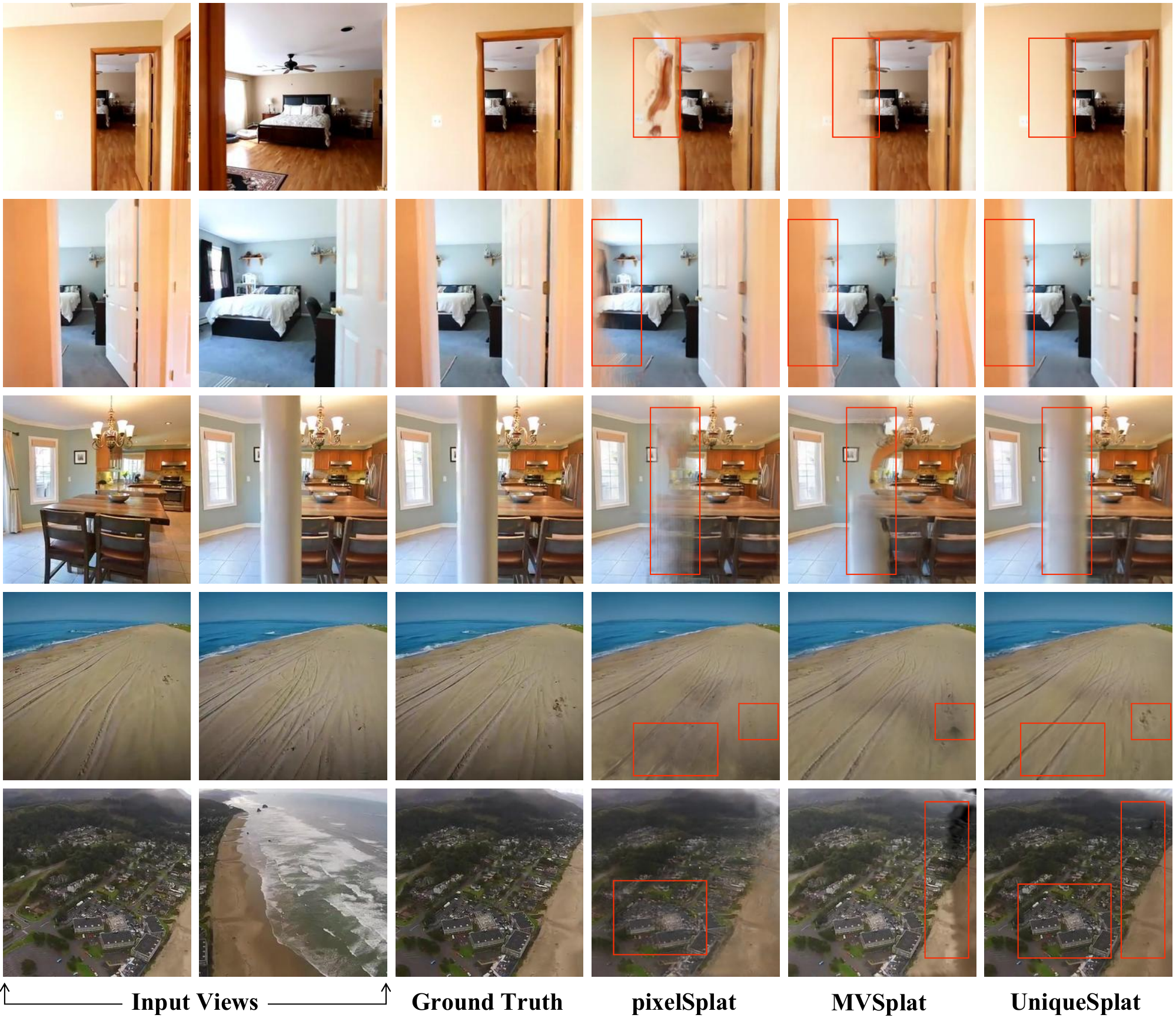}
    \caption{Qualitative comparison on RealEstate10K~\cite{zhou2018stereo} and ACID~\cite{liu2021infinite} benchmarks. All models are trained and tested on the same datasets. Benefiting from the information of target views, our method outperforms baseline models with higher rendering quality and better details.}
    \label{fig:intra}
\end{figure*}

\begin{figure*}
    \centering
    \includegraphics[width=\linewidth]{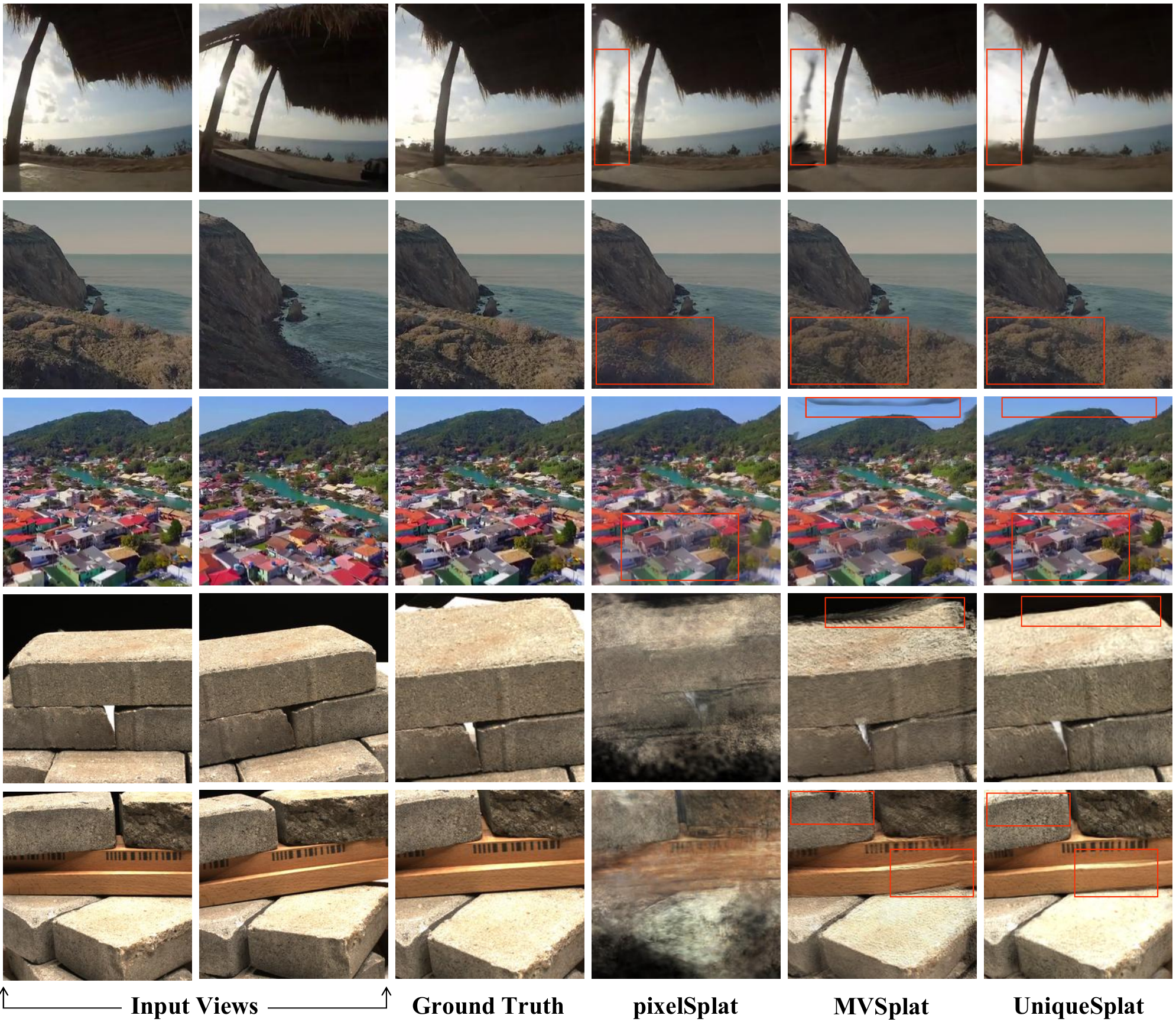}
    \caption{Qualitatitive comparison for corss-dataset generalization on ACID~\cite{liu2021infinite} and DTU~\cite{jensen2014large}. All models are trained on RealEstate10K~\cite{zhou2018stereo} and tested on ACID~\cite{liu2021infinite} (first three row) and DTU~\cite{jensen2014large} (last two rows) respectively.}
    \label{fig:cross}
\end{figure*}

In this section, we evaluate the effectiveness of our proposed framework on popular benchmark datasets. 
We first introduce the datasets, the evaluation metrics for novel view synthesis and implementation details in our experiment. 
We then present comprehensive experimental results, comparing our method with other state-of-the-art methods both quantitatively and qualitatively.
Finally, we perform ablation studies to evaluate the effects of our proposed components in our framework.

\subsection{Settings}
\textbf{Datasets.} 
To comprehensively evaluate the effectiveness of our method, we conduct extensive experiments on two large-scale datasets: RealEstate10K~\cite{zhou2018stereo} and ACID~\cite{liu2021infinite}. The RealEstate10K dataset consists of video frames of real estate scenes, with 67,477 scenes for training and 7,289 scenes for testing.
The ACID dataset comprises natural landscape scenes, divided into 11,075 training scenes and 1,972 testing scenes.
Both datasets provide estimated camera intrinsic and extrinsic parameters for each frame.
Following the setting in pixelSplat~\cite{charatan2024pixelsplat}, we evaluate all methods on three target novel viewpoints per test scene.
To further demonstrate the cross-dataset generalization ability of our approach, we additionally evaluate on the multi-view DTU~\cite{jensen2014large} dataset, which consists of object-centric scenes with corresponding camera poses. On DTU dataset, we report results on 16 validation scenes, with 4 novel views for each scene.

\textbf{Evaluation Metrics.} We evaluate the quality of rendered images with three metrics: pixel level PSNR, patch level SSIM~\cite{wang2004image}, and feature-level LPIPS~\cite{zhang2018unreasonable}. Higher PSNR and SSIM indicate better image quality while lower LPIPS correspond to higher perceptual fidelity. 
For fair comparison, we follow existing models~\cite{charatan2024pixelsplat,chen2024mvsplat,zhang2025transplat} and conduct all experiments at a resolution of $256 \times 256$.

\textbf{Architectural Details.}
For the primary network, the encoder consists of a shallow ResNet-like CNN with 6 residual blocks, followed by a Transformer composed of 6 stacked Transformer blocks. The depth predictor employs a 2D U-Net implementation from~\cite{rombach2022high} to refine the cost volumes, followed by a convolutional block for depth prediction and an additional 2D U-Net for further depth refinement.
For the view-specific branch, a ResNet18 network is first applied to extract features from the input views. After fusing these features with the target view matrix, 4 convolutional blocks and a linear layer map the fused features into a view-specific embedding. The view-specific embedding has the same dimensionality as the view-agnostic embedding. Both hypernetworks share the same architecture, a MLP network with 3 layers to map the embedding into the convolutional weights.

\textbf{Implementation Details.} 
Our implementation is based on the PyTorch framework with an off-the-shelf 3DGS render implemented in CUDA. All experiments are conducted on eight A6000 GPUs, utilizing the Adam optimizer with a learning rate of \(1 \times 10^{-6}\).
To maintain consistency with prior work, the model is trained with two input views and four target views per scene. For all the RealEstate10K, ACID and DTU datasets, all experiments are conducted at a resolution of \(256 \times 256\).
The encoder and predictor are initialized with those of MVSplat. In the view-conditioned network, the embedding dimension is set to 512. A ResNet backbone is employed to extract per-view features in the view-specific branch, producing feature maps of size \(256 \times 256\). These features are then projected and downsampled to generate the weights for the convolutional layers in the predictor.
The training loss is a weighted combination of Mean Squared Error (MSE) and Learned Perceptual Image Patch Similarity (LPIPS) losses. MSE is applied throughout the training process, while LPIPS is introduced after 150,000 iterations. The respective loss weights are set to 1 for MSE and 0.05 for LPIPS.

\subsection{Quantitative Evaluation}
In this section, we present quantitative results comparing our method with current state-of-the-art approaches on the RealEstate10K~\cite{zhou2018stereo}, ACID~\cite{liu2021infinite} and DTU~\cite{jensen2014large} datasets.
We first evaluate the performance of our method against several representative feed-forward methods~\cite{yu2021pixelnerf, suhail2022generalizable, du2023learning,xu2024murf} that focus on novel view synthesis from sparse views in the intra-test setting. GPNR~\cite{suhail2022generalizable} and AttnRend~\cite{du2023learning} are based on Light Field networks, while pixelNeRF~\cite{yu2021pixelnerf} and MuRF~\cite{xu2024murf} are constructed on NeRF. Additionally, we also compare with the state-of-the-art 3DGS-based models~\cite{charatan2024pixelsplat, chen2024mvsplat, zhang2025transplat}.
We present quantitative results on the RealEstate10K and ACID datasets in Table~\ref{tab:intratest}, where our method outperforms all previous state-of-the-art approaches across all metrics, achieving higher rendering quality for both indoor and outdoor scenes. 
Specifically, our method applies MVSplat~\cite{chen2024mvsplat} as the primary network, and our proposed view-conditioned hypernetwork incorporates view-dependent knowledge, resulting in a 0.89 dB increase in PSNR compared to MVSplat, reaching a PSNR of 27.28 dB on the RealEstate10K benchmark.
Even when compared to the latest method, TranSplat~\cite{zhang2025transplat}, our approach achieves a notable improvement, surpassing TranSplat by 0.59 dB in PSNR. Furthermore, our method reduces the LPIPS score to 0.119, demonstrating a significant enhancement in perceptual image quality. 
On the ACID dataset, our method also achieves state-of-the-art results, with a PSNR increase to 29.06 dB and an SSIM of 0.855. The LPIPS score is reduced to 0.134, further highlighting our model’s ability to generate realistic and perceptually consistent views across diverse scenes.
To evaluate the generalization ability of UniqueSplat, we conduct cross-dataset experiments, where all methods are trained on the RealEstate10K dataset and tested on other datasets. As shown in Table~\ref{tab:crosstest}, our method achieves superior performance on out-of-distribution novel scenes. This is due to the view-conditioned hypernetwork, which learns view-dependent information as prior knowledge, enabling the customization of 3D radiance fields.
For instance, when trained on the RealEstate10K dataset, our method achieves a PSNR of 28.63 dB, outperforming the current state-of-the-art method, TranSplat~\cite{zhang2025transplat}, by 0.46 dB when tested on the ACID dataset. Notably, our cross-dataset generalization results on ACID even surpass those of MVSplat, which was specifically trained on the ACID dataset.
Additionally, we perform experiments in a zero-shot setting on an object-centric dataset. We train our model on the RealEstate10K dataset and directly test it on the DTU dataset. As shown in Table~\ref{tab:crosstest}, UniqueSplat exhibits a significant improvement in generalization compared to previous feed-forward 3D Gaussian Splatting methods, which predict fixed Gaussians. 
This suggests that incorporating view-conditioned knowledge also enhances generalization ability on object-centric dataset.
Our results indicate that our method performs well when exposed to views from unseen scenes. While other methods experience a drop in generalization ability due to the domain gap between datasets, our approach is less affected, demonstrating the benefit of the view-conditioned branch for improved generalizability.
Overall, our method demonstrates strong performance in both intra-dataset and cross-dataset evaluations, achieving superior results compared with state-of-the-art methods.
\begin{figure*}
    \centering
    \includegraphics[width=1\textwidth]{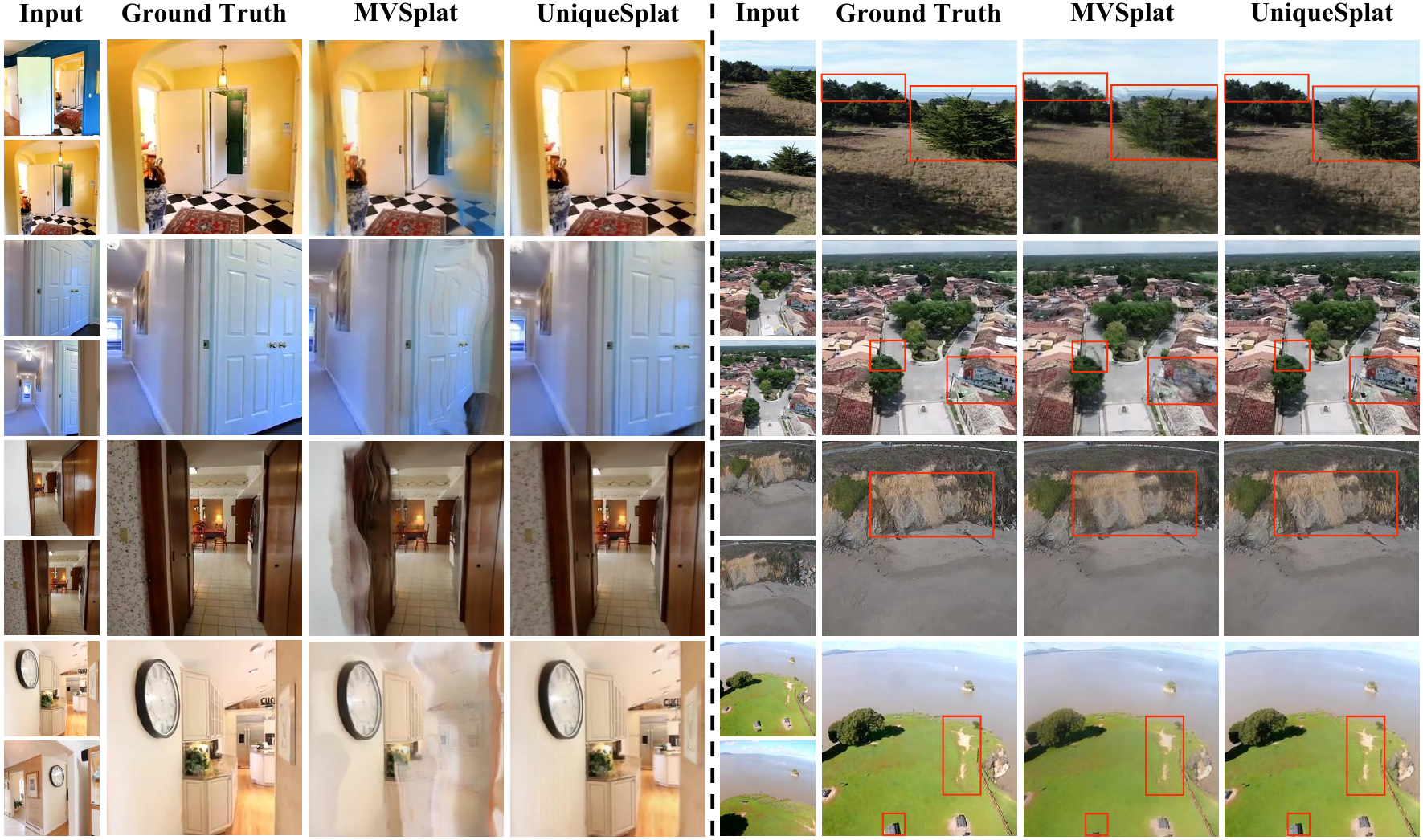}
    \caption{More qualitative comparisons with MVSplat. We train models on RealEstate10K dataset and evaluate them on both RealEstate10K and ACID datasets.}
    \label{fig:vis_sup}
\end{figure*}

\subsection{Qualitative Evaluation}
In this section, we present qualitative comparisons with pixelSplat~\cite{charatan2024pixelsplat} and MVSplat~\cite{chen2024mvsplat}.
We first present the intra-test results in Figure~\ref{fig:intra}, with the first three rows corresponding to the RealEstate10K dataset and the last two rows representing the ACID dataset. PixelSplat and MVSplat exhibit cracks and distortions at large viewing angles, while our method produces high-fidelity, realistic images, benefiting from the view-conditioned information. The first row demonstrates a scene with a large angular transformation, where our method accurately reconstructs the room, while pixelSplat and MVSplat show significant distortions. Additionally, our method excels in reconstructing scene structures, as shown in the third scene, where it successfully reconstructs a column, an area where the other methods struggle. Finally, in outdoor scenes, our method outperforms both pixelSplat and MVSplat, which exhibit blurriness in the last two rows.
We further visualize the rendering results in Figure~\ref{fig:cross}, where we train our model on the RealEstate10K dataset and test it on the ACID (first three rows) and DTU (last two rows) datasets. The dataset gap leads to blurry images and artifacts in pixelSplat and MVSplat. In contrast, UniqueSplat leverages additional target view information to render high-quality novel views with less edge blurriness, fewer artifacts, and more accurate spatial alignment, demonstrating superior generalization ability.
Since MVSplat serves as our primary network, we provide additional visualizations to demonstrate the effectiveness of incorporating view-conditioned knowledge. We train our models on the RealEstate10K dataset and evaluate them on both the RealEstate10K and ACID datasets. As illustrated in Figure~\ref{fig:vis_sup}, our proposed UniqueSplat synthesizes novel view with better structure preservation, more intricate details and fewer blurs.

\subsection{Efficiency Analysis}
In this section, we provide a detailed comparison of our method with existing approaches in terms of computational resources required for inference. Specifically, we conduct a comprehensive evaluation of all methods on an A6000 GPU, reporting the number of parameters, inference time, memory usage, and PSNR performance on the Re10K, ACID, and DTU datasets.

As shown in Table~\ref{tab:statistics}, PixelSplat contains 125.4M parameters, with an inference time of 0.282s and 4.19 GB of memory usage. These costs are primarily due to its epipolar transformer. MVSplat leverages a cost-volume encoder, achieving superior quantitative results with a lightweight model size and fast speed. In particular, MVSplat requires only 12.0M parameters, an inference time of 0.048s, and 1.68 GB memory usage, while outperforming PixelSplat by 0.5, 0.11, and 1.05 in PSNR on the three datasets, respectively.

TransSplat~\cite{zhang2025transplat} builds upon MVSplat by introducing a transformer-based module to produce high-confidence depth estimates and incorporating monocular depth priors. While this design provides modest PSNR improvements, it comes with higher computational costs, including 110.5M parameters, an inference time of 0.083s, and 4.98 GB memory.

In contrast, our method also extends MVSplat but leverages a two-branch view-conditioned hypernetwork to reconstruct customized Gaussians for each query view. This design achieves superior PSNR performance (27.28, 29.06, and 16.01 on Re10K, ACID, and DTU, respectively) while requiring lower computational resources (38.1M parameters, an inference time of 0.065s, and 3.12 GB memory) compared with TransSplat, demonstrating a more efficient trade-off between performance and cost.

\begin{table*}[ht]
    \centering
    \caption{Comparison of computational resources among different feed-forward Gaussian Splatting methods. We report the number of parameters (M), inference time (s), and memory usage (GB), along with PSNR performance on the Re10K~\cite{zhou2018stereo}, ACID~\cite{liu2021infinite}, and DTU~\cite{jensen2014large} datasets.}
    \label{tab:statistics}
    \normalsize
    \begin{tabular}{ccccccc}
        \toprule
        \multirow{2}{*}{Method} & \multirow{2}{*}{Params (M)}& \multirow{2}{*}{Time (s)} &  \multirow{2}{*}{Memory (GB)} & \multicolumn{3}{c}{PSNR} \\
        \cmidrule(lr){5-7}
        & & & & Re10K~\cite{zhou2018stereo} & ACID~\cite{liu2021infinite} & DTU~\cite{jensen2014large} \\
        \cmidrule(lr){1-7}
        pixelSplat~\cite{charatan2024pixelsplat} &125.4  &0.282  & 4.19 &25.89 &28.14 &12.89\\
        MVSplat~\cite{chen2024mvsplat}    &12.0  & 0.048 & 1.68 &26.39 & 28.25&13.94\\
        TransSplat~\cite{zhang2025transplat} &110.5   &0.083   & 4.98 &26.69 & 28.35&14.93\\
        Ours     & 38.1  &0.065   & 3.12 & 27.28 &29.06 & 16.01\\
        \bottomrule
    \end{tabular}
\end{table*}

\begin{table*}[t]
    \centering
    \caption{Ablation studies on the RealEstate10K~\cite{zhou2018stereo} and ACID~\cite{liu2021infinite} datasets. VA represents our view-agnostic branch and VS is our view-specific branch. We report inference time as well as the average PSNR, SSIM, and LPIPS on all test scenes.}
    \normalsize
    \setlength{\tabcolsep}{4mm}{\begin{tabular}{cccccccccc}
        \toprule
        \multirow{2}{*}{ID} &\multirow{2}{*}{VA} &\multirow{2}{*}{VS} &\multirow{2}{*}{Time (ms)} & \multicolumn{3}{c}{RealEstate10K~\cite{zhou2018stereo}} & \multicolumn{3}{c}{ACID~\cite{liu2021infinite}} \\
        \cmidrule(lr){5-7} \cmidrule(lr){8-10}
        & & & & PSNR↑ & SSIM↑ & LPIPS↓ & PSNR↑ & SSIM↑ & LPIPS↓ \\
        \cmidrule(lr){1-10}
        (a) &- &-            &48.5 & 26.39 & 0.869 & 0.128 & 28.25 & 0.843 & 0.144 \\
        (b) &$\checkmark$ &- &48.6 & 26.70 & 0.873 & 0.124 & 28.70 & 0.851 & 0.139 \\
        Ours &$\checkmark$ &$\checkmark$ &65.2 & \textbf{27.28} & \textbf{0.881} & \textbf{0.119} & \textbf{29.06} & \textbf{0.855} & \textbf{0.134} \\
        \bottomrule
    \end{tabular}}
    \label{tab:ablation_efficiency}
\end{table*}

\subsection{Ablation Study}

In this section, we conduct a series of ablation studies to evaluate the effectiveness of the proposed view-conditioned hypernetworks in our framework, as shown in Table~\ref{tab:ablation_efficiency}. We start by introducing variant (a), a baseline model that follows a standard view synthesis pipeline. This model consists of a primary network, which includes a backbone encoder and a predictor, without additional view-conditioned components.
Next, we introduce variant (b), which incorporates a view-agnostic branch (VA) to abstract general scene knowledge across various views from multiple scenes. 
This branch aims to capture shared features that are not specific to any single view, thus contributing to a more robust and stable training process.
Benefiting from the shareable information across diverse views, the model retains a comprehensive understanding of scenes, leading to an increase of 0.31dB and 0.45dB on PSNR, compared to the vanilla model on RealEstate10K and ACID benchmarks, respectively. The view-agnostic branch introduces only 0.1 ms additional latency, since it employs an MLP module to generate weights.
Our view-specific branch is proposed to dynamically adjust the predictor to provide the additional geometry information for the target views, via combining features of input views for each target view query to extract view-specific knowledge. 
By integrating both view-specific and view-agnostic knowledge, our UniqueSplat achieves the best PSNR of 27.28dB and 29.06dB on both RealEstate10K and ACID benchmarks. The view-specific branch introduces approximately 16.6 ms additional latency, as it relies on a ResNet18 backbone to extract features from two input views and then projects them into the target view matrix. This projection step is the primary contributor to the increased inference latency.
The quantitative results demonstrate the efficacy of the proposed view-conditioned hypernetwork. The integration of both the view-agnostic and view-specific branches leads to superior performance across multiple benchmarks, confirming the importance of incorporating both types of scene knowledge for robust and high-quality view synthesis.

\subsection{Analysis}
\begin{itemize}
    \item Our UniqueSplat demonstrates strong performance on widely-used datasets. Unlike existing feed-forward 3D Gaussian Splatting models, which predict fixed Gaussians for each scene, UniqueSplat customizes Gaussians for each view query by incorporating view-conditioned knowledge.
    \item The proposed view-conditioned hypernetwork injects view-conditioned knowledge into the primary network via generating the parameters of primary network.  Compared to other state-of-the-art methods, the design of view-conditioned hypernetwork achieves superior performance in intra-test and cross-test settings. Qualitative results further demonstrate that our method synthesizes high-quality images with fewer cracks, blurs and artifacts. 

    \item Evaluations of both the view-agnostic and view-specific branches in the view-conditioned hypernetwork show that the introduction of both branches encouragingly improves the performance of novel view synthesis. The view-agnostic branch abstracts the characteristics shared across various views of the training scenes, contributing to a robust and stable training process. The view-specific branch introduces multi-view priors, dynamically adjusting the model to accommodate each specific view.

\end{itemize}

\begin{figure}[t]
    \centering
    \includegraphics[width=0.5\textwidth]{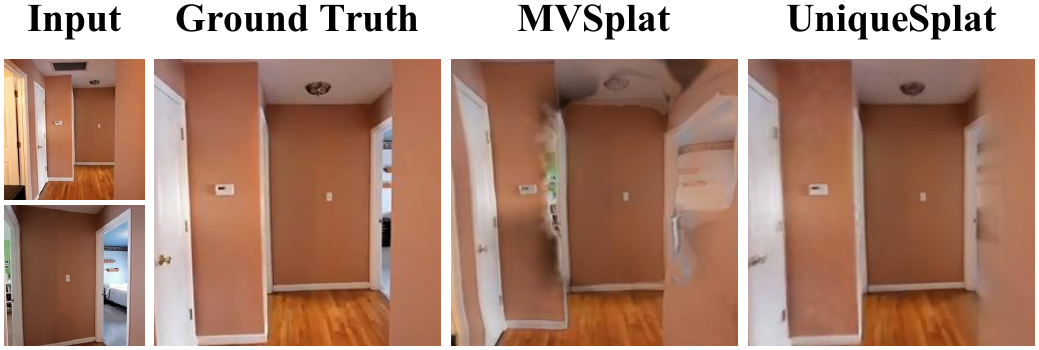}
    \caption{Failure case. While UniqueSplat delivers impressive results, it fails to accurately reconstruct unseen regions}
    \label{fig:fail_sup}
\end{figure}

\section{Limitations and Future Work}

While UniqueSplat achieves compelling results and outperforms previous methods, it still has certain limitations. Specifically, because it predicts pixel-aligned Gaussians for each input view, UniqueSplat may introduce artifacts when rendering unseen regions, as shown in Figure~\ref{fig:fail_sup}. 
Future work could address these issues by introducing generative models, such as diffusion models.

\section{Conclusion}
\label{sec:conclusion}

In this paper, we have proposed UniqueSplat, a novel view-conditioned 3D Gaussian Splatting model that reconstructs customized 3D radiance fields for each view query. 
Conventional feed-forward methods directly utilize fixed Gaussians to render images across all views, struggling to specialize to specific view.
To address this, UniqueSplat dynamically adjusts Gaussians in accordance with different views by learning the view-conditioned knowledge as a prior.
It simultaneously learns the view-agnostic embeddings and view-specific knowledge, incorporating shareable knowledge from various view of 3D scenes and adapting the Gaussians dynamically according to the given views.
Extensive experiments on benchmark datasets, including RealEstate10K, ACID and DTU, demonstrate that UniqueSplat outperforms state-of-the-art methods, both in terms of image quality and generalization ability.

\section*{Acknowledgments}
This work was supported in part by the National Natural Science Foundation of China under Grant 62576185, and by the Beijing Natural Science Foundation under Grant L252011.

\bibliographystyle{IEEEtran}
\bibliography{main}

\begin{IEEEbiography}[{\includegraphics[width=1in,height=1.25in,clip,keepaspectratio]{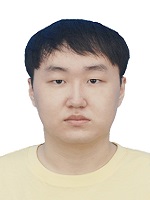}}]{Haixu Song}
	received the B.Eng. degree in Automation and the M.Eng. degree in Control Science and Engineering from the Harbin Institute of Technology in 2020 and 2022, respectively. He is currently pursuing the Ph.D. degree with the Department of Electronic Engineering, Tsinghua University. His main interest is 3D computer vision.
\end{IEEEbiography}


\begin{IEEEbiography}[{\includegraphics[width=1in,height=1.25in,clip,keepaspectratio]{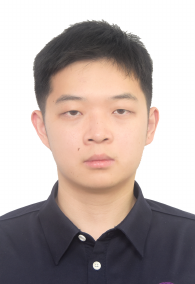}}]{Xiaoke Yang} received the B.Eng. degree from the School of Computer Science, Beijing University of Posts and Telecommunications, in 2022. He is currently pursuing the M.Eng. degree with the Tsinghua Shenzhen International School, Tsinghua University. His research interests include 3D deep learning and Deepfake Detection.
\end{IEEEbiography}

\begin{IEEEbiography}[{\includegraphics[width=1in,height=1.25in,clip,keepaspectratio]{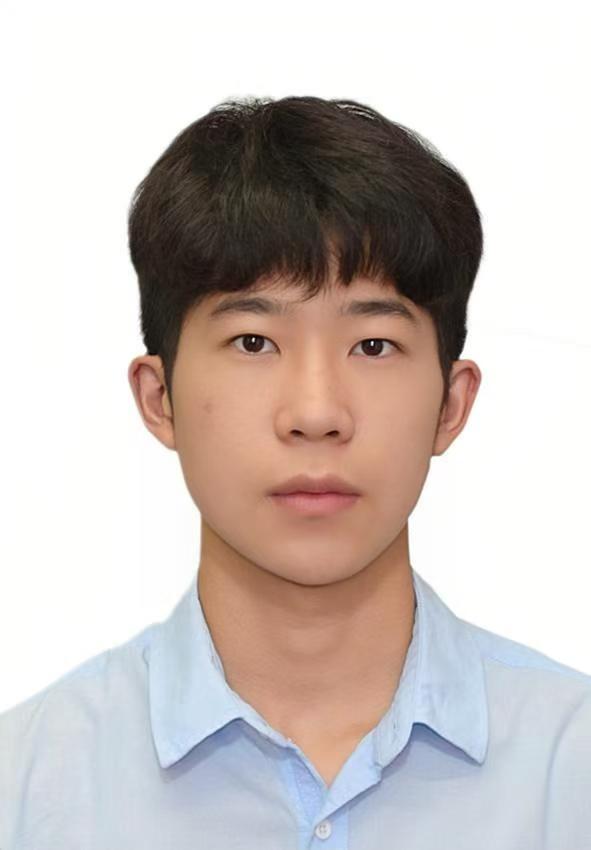}}]{Shengjun Zhang} received the B.Eng degree from the Department of Engineering Physics of Tsinghua University, Beijing, China, in 2023, where he is currently pursuing the Ph.D  degree with the Department of Electronic Engineering. His main research interest is 3D computer vision.
\end{IEEEbiography}

\begin{IEEEbiography}[{\includegraphics[width=1in,height=1.25in,clip,keepaspectratio]{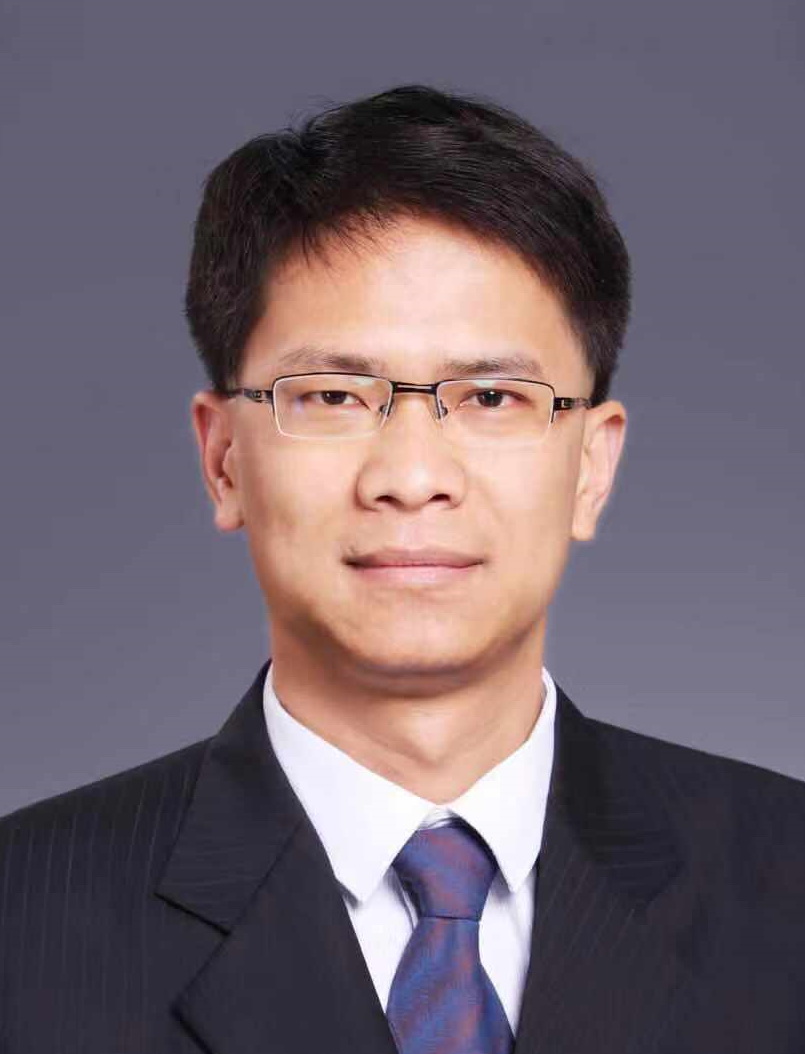}}]{Jiwen Lu}
	(Fellow, IEEE) received the B.Eng. degree in mechanical engineering and the M.Eng. degree in electrical engineering from the Xi’an University of Technology, Xi’an, China, in 2003 and 2006, respectively, and the Ph.D. degree in electrical engineering from Nanyang Technological University, Singapore, in 2012. He is currently a Professor with the Department of Automation, Tsinghua University, Beijing, China. His current research interests include computer vision and pattern recognition, where he has authored/coauthored more than 160 scientific articles in IEEE TRANSACTIONS ON PATTERN ANALYSIS AND MACHINE INTELLIGENCE, International Journal of Computer Vision, CVPR, ICCV, and ECCV. He serves as the Co-Editor-in-Chief for Pattern Recognition Letters and an Associate Editor for IEEE TRANSACTIONS ON IMAGE PROCESSING, IEEE TRANSACTIONS ON CIRCUITS AND SYSTEMS FOR VIDEO TECHNOLOGY, IEEE TRANSACTIONS ON BIOMETRICS, BEHAVIOR, IDENTITY SCIENCE, and Pattern Recognition. He also serves as the Program Co-Chair for FG’2023, VCIP’2022, AVSS’2021, and ICME’2020. He is a recipient of the National Outstanding Youth Foundation of China Award and an IAPR Fellow. 
\end{IEEEbiography}

\begin{IEEEbiography}[{\includegraphics[width=1in,height=1.25in,clip,keepaspectratio]{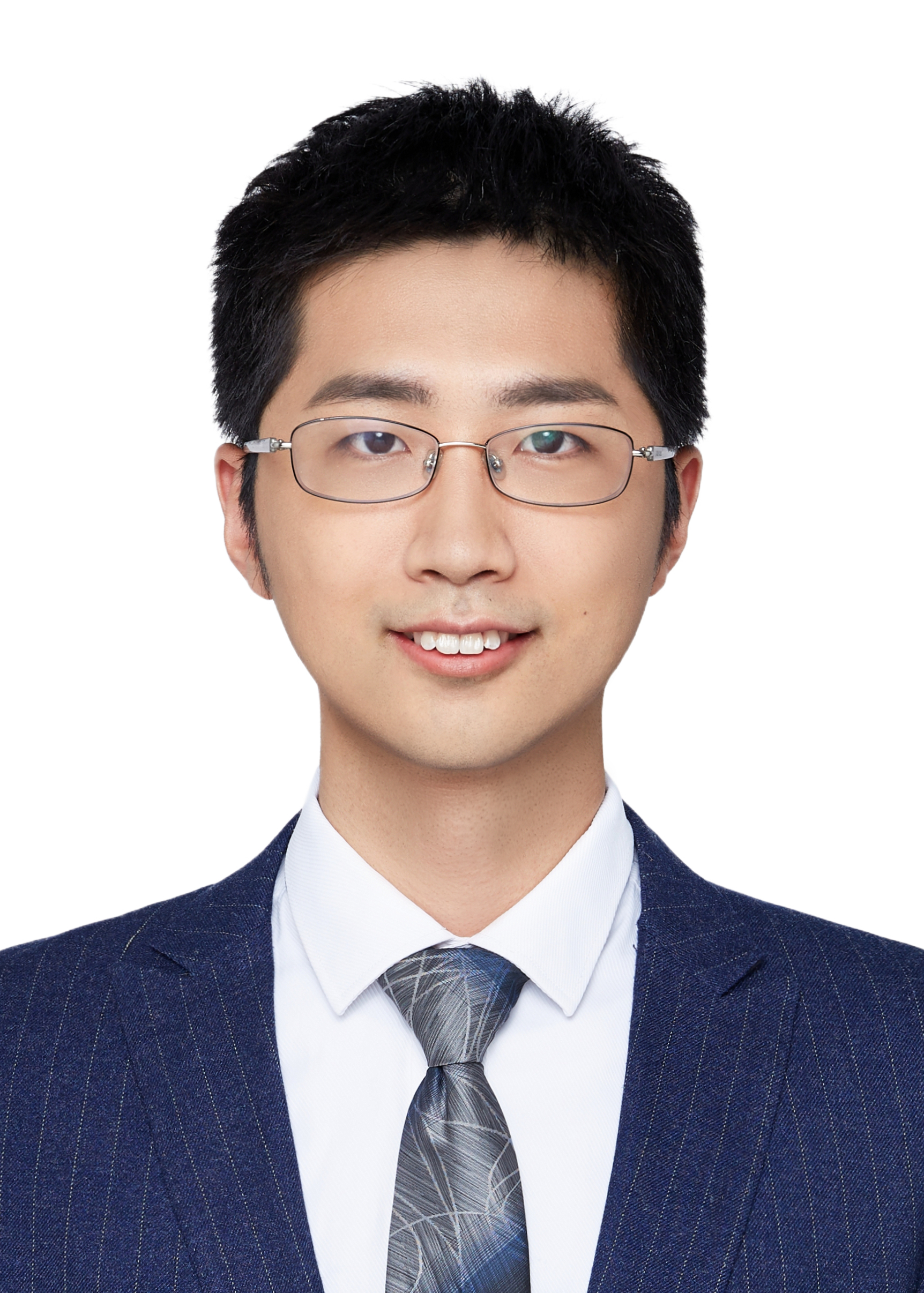}}]{Yueqi Duan}
	(Member, IEEE) received the B.S. and Ph.D. degrees from the Department of Automation, Tsinghua University, in 2014 and 2019, respectively. From 2019 to 2021, he was a Post-Doctoral Researcher with the Computer Science Department, Stanford University. He is currently an Assistant Professor with the Department of Electronic Engineering, Tsinghua University. He has published more than 30 scientific papers in the top journals and conferences, including IEEE TRANSACTIONS ON PATTERN ANALYSIS AND MACHINE INTELLIGENCE, IEEE TRANSACTIONS ON IMAGE PROCESSING, CVPR, ICCV, ECCV and NeurIPS. His research interests include computer vision and pattern recognition. He served as the Publication Chair for FG, the Area Chair for CVPR, ICLR, MM and ICME, and a Regular Reviewer for a number of journals and conferences, e.g., IEEE TRANSACTIONS ON PATTERN ANALYSIS AND MACHINE INTELLIGENCE, IEEE TRANSACTIONS ON IMAGE PROCESSING, IJCV, CVPR, ICCV, ECCV, ICML, NeurIPS, and SIGGRAPH. He was awarded the Excellent Doctoral Dissertation of Chinese Association for Artificial Intelligence (CAAI) in 2020.
\end{IEEEbiography}
\end{document}